\documentclass[conference,a4paper]{IEEEtran}
\IEEEoverridecommandlockouts

\usepackage{cite}
\usepackage{amsmath,amssymb,amsfonts}
\usepackage{graphicx}
\usepackage{booktabs}
\usepackage{adjustbox}
\usepackage{textcomp}
\usepackage{xcolor}
\usepackage{caption}
\usepackage{subcaption}
\usepackage{cuted}   
\usepackage{multirow}
\usepackage{hyperref}
\hypersetup{hidelinks}
\usepackage[capitalize]{cleveref}

\newcommand{\sota}[1]{\begingroup\bfseries\boldmath #1\endgroup}
\newcommand{\sotapm}[2]{$\boldsymbol{#1 \pm #2}$}
\newcommand{\ind}{\mathbf{1}}

\title{BIM-Native Tokenization for\\Constraint-Aware Room Layout Synthesis}

\author{%
\IEEEauthorblockN{Manuel Ladron de Guevara\IEEEauthorrefmark{1},
Jinmo Rhee\IEEEauthorrefmark{2},
Ardavan Bidgoli\IEEEauthorrefmark{1},
Vaidas Razgaitis\IEEEauthorrefmark{1},
Michael Bergin\IEEEauthorrefmark{1}}
\IEEEauthorblockA{\IEEEauthorrefmark{1}Higharc, United States \qquad
\IEEEauthorrefmark{2}University of Calgary, Canada\\
\{manuelrodriguez,\,ardavanbidgoli,\,vaidasrazgaitis,\,michaelbergin\}@higharc.com
\qquad jinmo.rhee@ucalgary.ca\\
\vspace{0.15em}
Project page: \url{https://manuelladron.github.io/sbm}}
}

\makeatletter
\def\ps@IEEEtitlepagestyle{%
  \def\@oddfoot{\mycopyrightnotice}%
  \def\@evenfoot{}%
}
\def\mycopyrightnotice{%
  {\footnotesize 979-8-3195-3697-6/26/\$31.00~\copyright~2026 IEEE\hfill}%
  \gdef\mycopyrightnotice{}%
}
\makeatother

\begin{document}
\maketitle

\begin{strip}
\centering
\includegraphics[width=0.74\textwidth]{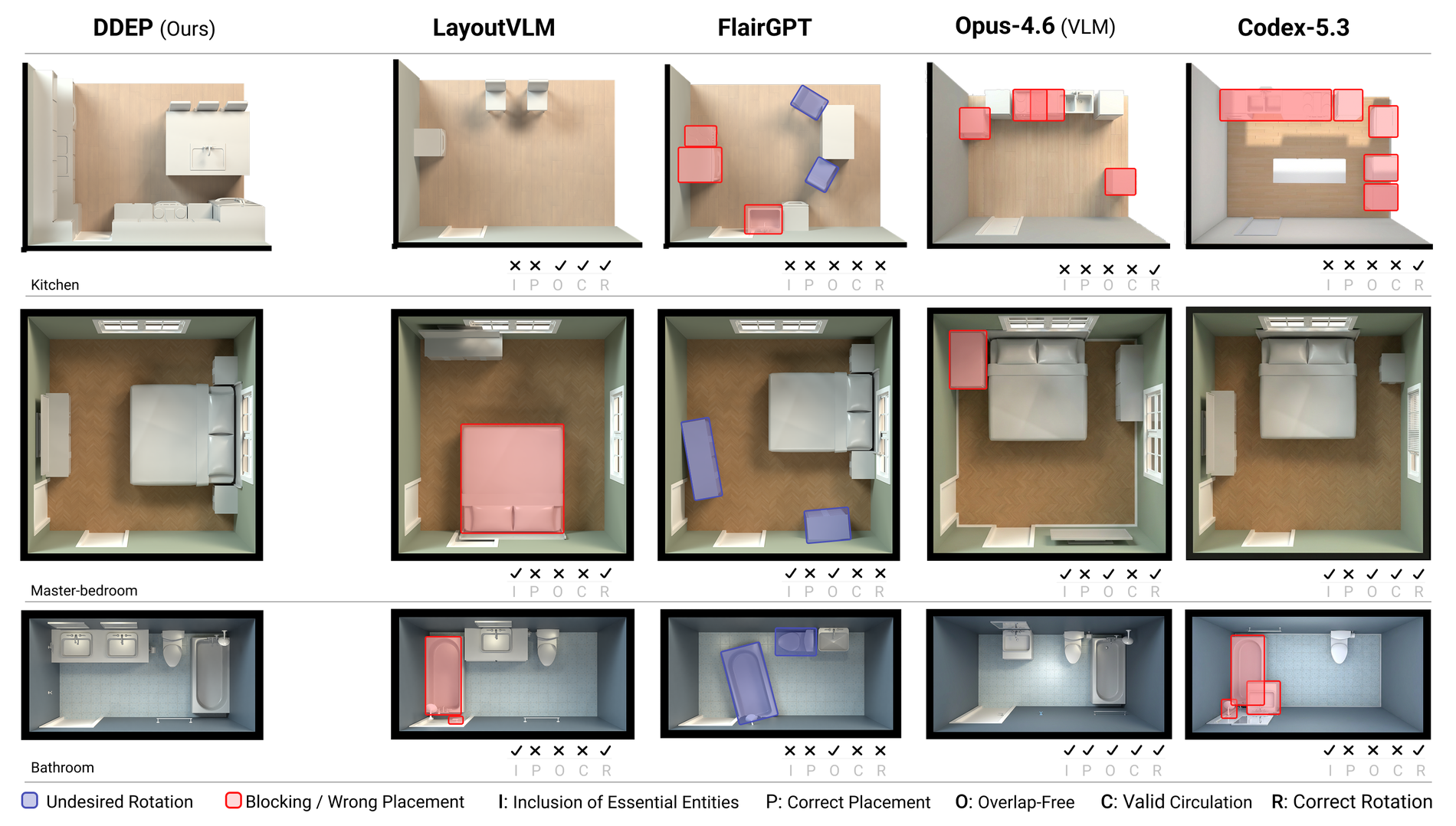}
\captionof{figure}{Qualitative comparison. Each row shows a different room type; columns show outputs from frontier LLM/VLM baselines, domain-specific methods, and our DDEP. DDEP layouts respect wall-referenced anchoring and door-connected circulation while satisfying the inventory program.}
\label{fig:hero}
\vspace{0.5em}
\end{strip}

\begin{abstract}
We present a BIM-native tokenization for room-level layout synthesis in Building Information Modeling (BIM) scenes. The core contribution is representational: we encode each room as a sequence of \emph{BIM-Token Bundles}, realized as columns of a sparse attribute--feature matrix that unifies categorical and continuous attributes of walls, openings, and entities under wall-referenced (translation/scale-invariant) coordinates. A mixed-type embedding module produces a unified token vector from this matrix; a single Transformer backbone is then trained in two modes: encoder-only for room embeddings and retrieval, and encoder--decoder for autoregressive entity placement, which we call Data-Driven Entity Prediction (DDEP). On a controlled same-data benchmark with shared ontology and evaluation harness, DDEP outperforms ATISS and BLT baselines bridged into our representation, with ablations identifying joint continuous-feature embedding and entity ordering as primary drivers. Encoder embeddings cluster rooms by type more tightly than large general-purpose text encoders, which in turn retain an edge on within-type ranking. We frame this work as evidence that modestly sized, domain-specific sequence models over well-designed BIM tokenizations are a useful primitive for constraint-aware spatial generation, complementary to general-purpose LLMs/VLMs which we also benchmark.
\end{abstract}

\begin{IEEEkeywords}
BIM, layout synthesis, structured tokenization, Transformer, constraint-aware generation.
\end{IEEEkeywords}

\section{Introduction}
\label{sec:intro}

Computer-Aided Design (CAD) and Building Information Modeling (BIM) scenes encode rich semantics, hierarchies, and domain constraints that go well beyond appearance. Most strong 3D generators---voxel, mesh, point-cloud, and image-conditioned diffusion models---treat scenes as unstructured geometry~\cite{eastman_spatial_1975}, yielding visually plausible but hard-to-edit outputs that often violate basic validity rules. Room-level layout design, in contrast, requires reasoning over semantics, references, and constraints at multiple scales, and producing \emph{parametric} outputs that downstream BIM tools can ingest and modify. Decades of computational approaches---from CAD macros and parametric families to rule systems, optimization, and learning-based methods---have aimed to automate this process~\cite{MONEDERO2000369,kan_automated_2017,Flager2007ACO}, yet progress remains bottlenecked less by the algorithm and more by the representation that the algorithm operates on. An effective representation must expose structure, generalize across typologies, and remain BIM-editable.

We argue that for room-level BIM layout this representation should be \emph{sequence-native}, \emph{mixed-type}, and \emph{wall-referenced}. Sequence-native, so a single Transformer backbone can serve retrieval and generation; mixed-type, so categorical (room type, family identifiers) and continuous (positions, sizes, rotations) attributes share one embedding pipeline; and wall-referenced, so spatial coordinates are invariant to absolute translation and scale and align with architectural reasoning. We instantiate these properties in a normalized tokenizer that emits \emph{BIM-Token Bundles}---sequence units encoding room topology, entity attributes, wall-referenced geometry, and relational structure---realized as columns of a sparse attribute--feature matrix, whose rows are token features, and fused via a mixed-type embedding module into a unified token vector. Unlike flat serializations, this matrix keeps every entity editable as a BIM object: an output token still carries its category, hosting wall, parametric position, offset, dimensions, and rotation rather than only an image-space box. We then train a single Transformer backbone in two modes: encoder-only for room embeddings and retrieval, and encoder--decoder for autoregressive entity placement, which we call Data-Driven Entity Prediction (DDEP); \cref{fig:hero} shows representative outputs.

We make three contributions. (1) A BIM-native, wall-referenced tokenization of residential room layouts paired with a mixed-type embedding module that jointly handles categorical, scalar, and grouped-continuous features. (2) A single Transformer backbone reused across retrieval and generation, with constrained decoding for BIM validity at inference. (3) A two-track evaluation: a \emph{controlled same-data benchmark} that isolates architectural effects under matched data and budgets, plus a deployment-realistic comparison against frontier LLMs/VLMs and domain-specific layout generators. We frame the LLM/VLM comparison as a calibration baseline rather than a fair architectural comparison, and discuss its limitations explicitly (\cref{sec:experiments}). The aim of the paper is not to claim a new general Transformer architecture, but to show that a careful BIM-native tokenization is a useful primitive for constraint-aware indoor layout synthesis.

\section{Related Work}
\label{sec:related}

\paragraph{Rule-based and optimization-based layout}
Early furniture-layout systems encoded interior-design guidelines as cost functions or grammars and solved them with sampling or evolutionary search~\cite{merrell_interactive_2011,song_web3d-based_2019,kan_automated_2017}. These approaches embed human design knowledge and remain interpretable, but are bounded by their predefined rules and require manual rule authoring per typology.

\paragraph{Deep generative layout models}
Graph-based generators model layouts as scene graphs and predict node attributes and relational edges~\cite{xu2018graph2seqgraphsequencelearning,hu2020graph2plan}. Image-based approaches~\cite{tanasra_automation_2023,liu_exploration_2024} and diffusion models~\cite{nauata2021housegan,shabani2022housediffusion,midiffusion2024,semLayoutDiff2025,nguyen2024housecrafter,layyourscene2025,directlayout2025} learn strong priors over floorplans and 3D scenes. We differ from floorplan generators such as HouseDiffusion~\cite{shabani2022housediffusion}: our output is a sequence of \emph{wall-referenced parametric entities} hosted on room walls rather than a polygonal floorplan, so direct comparison requires output-format bridging (\cref{sec:experiments}).

\paragraph{Transformer-based indoor scene synthesis}
ATISS~\cite{Paschalidou2021NEURIPS} and SceneFormer~\cite{wang2020sceneformer} treat indoor scene synthesis as autoregressive generation of furniture objects with discretized/normalized geometric attributes, conditioned on a room boundary. CLIP-Layout~\cite{Liu2023CLIPLayoutSI} augments this with CLIP-based style embeddings. BLT~\cite{10.1007/978-3-031-19790-1_29} performs masked, iterative layout prediction for graphic design. These methods assume flat object vocabularies with absolute or grid-aligned poses; they do not preserve BIM's heterogeneous, wall-hosted, mixed categorical--continuous structure for downstream editing. We compare directly against ATISS and BLT under a shared harness, with the bridging caveat discussed in \cref{sec:experiments}.

\paragraph{LLM/VLM-based layout generation}
LayoutGPT~\cite{feng2023layoutgpt} casts layout synthesis as compositional planning over object--relation tokens; LayoutVLM~\cite{sun2025layoutvlm} couples vision and language features to optimize 3D scenes; HouseTune~\cite{zong2024housetune} and FlairGPT~\cite{littlefair2025flairgpt} use LLMs for floorplan refinement and stylistic exploration. LLM4CAD~\cite{li2024llm4cad} tokenizes CAD elements symbolically. FloorPlan-DeepSeek~\cite{yin2025floorplandeepseek} performs autoregressive next-room prediction at the floorplan level. None of these unify semantic, relational, and wall-referenced continuous attributes in a single sequence-native form. Our LLM/VLM benchmark (\cref{sec:experiments}) is best read as a deployment-realistic floor; we explicitly do not claim it is a fair architectural comparison, as fully-trained DDEP is matched against off-the-shelf zero-shot models.

\section{Method}
\label{sec:method}

\begin{figure*}[t]
\centering
\includegraphics[width=0.95\linewidth]{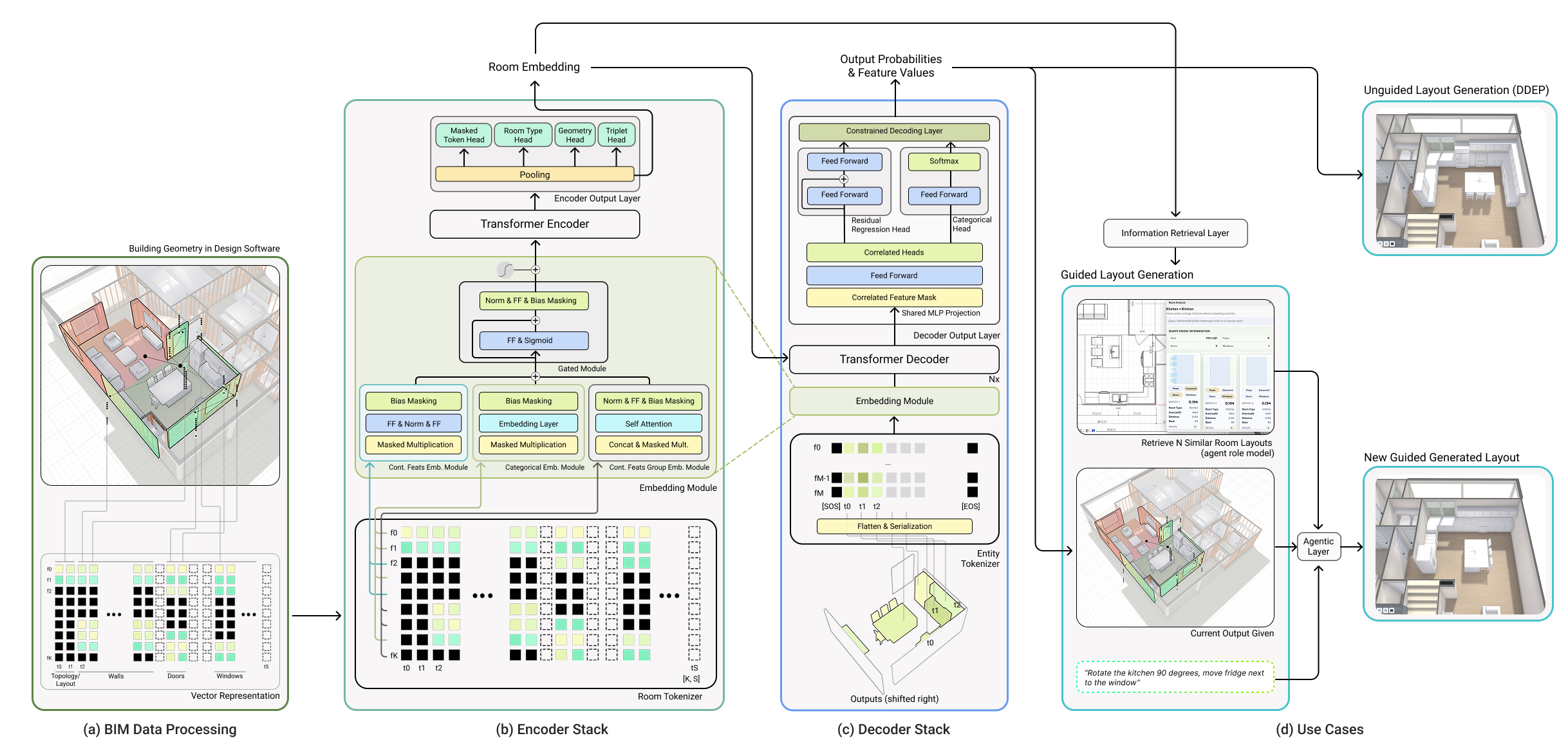}
\caption{Model overview. (a) BIM data extraction and assembly into a discrete set of BIM-Token Bundles. (b) The encoder stack processes the tokenized attribute--feature matrix and outputs a room representation. (c) The decoder stack consumes the room representation as memory for cross-attention and the entity sequence as input, trained with next-token prediction. (d) Downstream use cases. Retrieval (\cref{sec:embeddings}) and unguided generation (\cref{sec:controlled_results,sec:llm_floor}) are evaluated in this paper; the user-guided path through an agentic layer is a prospective integration that our BIM-native output tokens enable but that we do not evaluate here.}
\label{fig:overview}
\end{figure*}

\Cref{fig:overview} gives the pipeline. A room is first converted into typed tokens. The \emph{encoder} reads only the room envelope---walls, doors, windows---and compresses it into a contextual memory plus a pooled room embedding used for retrieval. The \emph{decoder} reads the sequence of room contents and, attending to that memory, predicts each next entity with its continuous placement parameters. Both modes share one tokenization and one set of weights, differing only in which sequence they read and which heads they use.

\subsection{Room Representation}
\label{sec:representation}
We operate on residential room-level layouts extracted from professional BIM projects, expressed in a local coordinate frame normalized for position and scale. Each room is decomposed into a \emph{room envelope} and \emph{room contents}:
$r_{\text{env}} = (y^{\text{topo}}, y^{\text{layout}}, \mathcal{E}, \mathcal{D}, \mathcal{W})$,
$r_{\text{ent}} = (\mathcal{P}, \mathcal{C})$.
Here $y^{\text{topo}}$ is a discrete room-type token, $y^{\text{layout}}$ aggregates global scalars (area, perimeter); $\mathcal{E}, \mathcal{D}, \mathcal{W}$ are walls, doors, and windows; $\mathcal{P}, \mathcal{C}$ are props and casework. The encoder consumes only the envelope, while the decoder generates the entity sequence.

\paragraph{Walls and openings}
Walls are $N_E$ directed segments $e_j = (x_j^{(1)}, x_j^{(2)}, a_j^{\text{wall}})$ whose counter-clockwise order defines the room polygon. Doors and windows are wall-hosted: each door $d_k = (j_k, t_k, w_k, a_k^{\text{door}})$ stores the supporting edge index, a normalized position $t_k \in [0,1]$ along that edge, opening width, and categorical attributes (family, swing). Windows are defined analogously. This parameterization is invariant to absolute translation and scale, and it preserves the BIM convention that many architectural elements are hosted by, and edited relative to, a wall.

\paragraph{Entities}
Each room-content entity $q \in \mathcal{P}\cup\mathcal{C}$ is parameterized as $q = (c_q, j_q, t_q, \delta_q, s_q, \rho_q, a_q^{\text{ent}})$, where $c_q$ is the entity type, $j_q$ is the supporting edge, $t_q$ is the wall-relative position, $\delta_q$ is the lateral offset from the wall, $s_q$ is size, $\rho_q$ is rotation (props only), and $a_q^{\text{ent}}$ collects further categorical attributes.

\subsection{BIM Tokenization and Mixed-Type Embedding}
\label{sec:tokenization}

\paragraph{BIM-Token Bundles}
We map each room to two sequences of \emph{BIM-Token Bundles}: an envelope sequence $(\tau^{\text{CLS}}, \tau^{\text{t}}, \tau^{\text{l}}, \tau^{\text{e}}_1,\dots,\tau^{\text{d}}_1,\dots,\tau^{\text{w}}_1,\dots,\tau^{\text{EOS}})$ for the encoder, and an entity sequence $(\tau^{\text{SOS}}, \tau^{\text{ent}}_1,\dots,\tau^{\text{EOS}})$ for the decoder. Each bundle corresponds to a single logical element (topology, layout, wall, door, window, or entity).

\paragraph{Attribute--feature matrix}
We realize these heterogeneous sequences as a sparse attribute--feature matrix $X \in \mathbb{R}^{F \times S}$, where each column $X_{:,s}$ encodes one bundle and each row corresponds to one feature (e.g., \textit{token\_type\_id}, \textit{token\_id}, edge endpoints, edge length, $t$-value, $\delta$-offset, size, rotation). Only a subset of features is active for any given token type; inactive entries are filled with a sentinel padding value $p{=}{-}100$. This differs from flat field serialization: a token remains a complete BIM element, and its active rows expose the typed attributes needed to edit that element later. This sparse layout lets the same backbone handle topology tokens, wall tokens, opening tokens, and entity tokens uniformly without forcing all element types into a lossy common schema. Encoder and decoder use disjoint feature sets via $X^{\text{enc}}$ and $X^{\text{dec}}$.

\paragraph{Mixed-type embedding}
A shared FeatureEmbedding module turns $X$ into dense token vectors. For each active feature $f$:
\textit{(i)} categorical features (type/family identifiers, edge indices) use learnable embedding tables;
\textit{(ii)} scalar continuous features (area, perimeter, length) use small MLPs;
\textit{(iii)} grouped continuous features (e.g.\ edge endpoint pairs, opening corner distances) use a \emph{MultiContinuousEmbedding} that embeds each scalar in the group and aggregates via self-attention. A valid-feature mask $m_{f,s} = \ind[X_{f,s} \neq p]$ zeros out contributions from inactive features (including the bias terms of any linear layers). The bundle embedding is the sum of all active feature embeddings,
$e_s = \sum_{f} m_{f,s}\, E_f(X_{f,s}) \in \mathbb{R}^d,$
optionally augmented with learned positional embeddings.

\subsection{Backbone and Operating Modes}
\label{sec:backbone}

We use a standard Transformer encoder--decoder backbone with two operating modes that share weights.

\paragraph{Encoder-only mode}
Given the envelope embeddings $E^{\text{enc}}$, the encoder produces memory $M=\mathrm{Enc}_\theta(E^{\text{enc}}) \in \mathbb{R}^{S_{\text{enc}}\times d}$. We pool at the CLS position to obtain a room embedding $z(r_{\text{env}})=M_0$, used for retrieval and clustering. Auxiliary heads (room-type classification, masked-token prediction) operate on $M$.

\paragraph{Encoder--decoder mode (DDEP)}
The encoder produces $M$ as above; the decoder consumes entity embeddings $E^{\text{dec}}$ and attends causally to its own prefix and via cross-attention to $M$, emitting $H=\mathrm{Dec}_\theta(E^{\text{dec}}, M)$. Per-token heads produce mixed categorical and continuous outputs (entity type, edge attachment, $t$, $\delta$, size, rotation). At inference, a constrained decoding layer filters the model's output distribution to enforce BIM validity, including egress clearance and door-swing zones. This layer is a validity guard over candidate tokens, not a post-hoc layout optimizer.

\paragraph{Training}
We use a two-stage schedule: \emph{(i)} encoder pretraining with a composite loss combining room-type classification, masked-token prediction, graded triplet contrastive learning, and geometric preservation; \emph{(ii)} encoder--decoder fine-tuning for DDEP via teacher forcing, with cross-entropy over categorical heads and MSE over continuous heads.

\section{Experiments}
\label{sec:experiments}

We evaluate the proposed tokenization and backbone on a corpus of professional residential BIM scenes (single-family homes with typed rooms).

\subsection{Setup, Splits, and Metrics}
\label{sec:setup}

We use two complementary protocols whose results are not directly comparable.

\textbf{Controlled same-data benchmark (lead).} A frozen split of $15{,}740/2{,}009/1{,}826$ rooms (train/val/test) across five room types (\texttt{bathfull}, \texttt{bedroom}, \texttt{laundry}, \texttt{living}, \texttt{master\_bed}) with shared ontology normalization, canonical furniture dimensions, and a unified evaluation harness. All methods are trained from scratch under a matched budget of 200 epochs with no method-specific tuning and evaluated on the shared 1{,}225-room geometry-valid intersection. This protocol is intentionally smaller than production DDEP; its purpose is to isolate architectural effects under matched data conditions rather than to report our best deployment configuration.

\textbf{Frontier LLM/VLM baseline (contextual).} Production DDEP follows a two-stage pipeline---encoder pre-training on the full augmented corpus ($439{,}932$ samples over 53 room types), then per-room-type encoder--decoder fine-tuning on the $75{,}720$-sample DDEP subset (14 types)---and is compared off-the-shelf against zero-shot frontier text and vision-language models (Claude Opus/Sonnet 4.6, Claude Haiku 4.5, GPT-5.2, Gemini 3.1 Pro, Codex 5.3, Qwen 3.5) and two domain-specific systems (LayoutVLM~\cite{sun2025layoutvlm}, FlairGPT~\cite{littlefair2025flairgpt}) on a 50-room held-out set. This is a deployment-realistic floor rather than the paper's headline architectural claim; we discuss its limits in \cref{sec:llm_floor}. Extended implementation details, per-room-type breakdowns, and further qualitative examples are available on the project page.

\paragraph{Metrics}
We report metrics under identical geometry and inventory specifications. The definitions below are stated over our full room ontology, since the frontier protocol spans ten room types; the controlled protocol exercises the five listed above. \textbf{Coverage} (Cov., $\uparrow$) is a semantic inventory score. For category $i$, predicted count $p_i$ is compared with the allowed range $[m_i^{\min},m_i^{\max}]$ using missing and overfill fractions $d_i=\max(0,m_i^{\min}-p_i)/\max(1,m_i^{\min})$ and $o_i=\max(0,p_i-m_i^{\max})/\max(1,m_i^{\max})$. The item score is $s_i=1-\mathrm{clip}(d_i+0.5\,o_i,0,1)$, and room coverage averages weighted item scores ($1.0$ for essential items, $0.5$ for optional items) with alternative requirement groups, such as tub-or-shower in full baths or range versus oven-plus-cooktop in kitchens. A capped unsupported-extra penalty discourages hallucinated entities. Thus Cov.\ measures \emph{what} was placed, not whether it is reachable or collision-free.

\textbf{Navigability} evaluates functional access from inward-offset door portals to essential targets on a clearance-inflated walkable grid. Beds contribute two side-access targets; sinks, toilets, ranges, and dressers contribute front-access targets. \textbf{SR} ($\uparrow$) is the fraction of door--target pairs reachable by A* routing. \textbf{DF} ($\downarrow$) averages the capped detour penalty $f(\rho)=\min(1,(\rho-1)/2)$ over reachable pairs, where $\rho$ is path length over Euclidean distance; if no pair is reachable, DF is set to $1$. We report $\mathrm{Nav}=100(\mathrm{SR}-0.35\,\mathrm{DF})$, so severely blocked rooms can score below zero.

\textbf{Overlap--Clearance} (OC, $\downarrow$) uses exact footprint polygons: $\mathrm{OC}=100(0.5\,\mathrm{EOF}+0.2\,\mathrm{GOA}+0.3\,\mathrm{DCI})$, where EOF is per-entity overlap with eligible neighbors, GOA is total multiply occupied floor area normalized by room area, and DCI is door-clearance intrusion. Eligible overlaps include prop--prop and same-category casework collisions, while prop--casework overlaps are ignored because they often encode intentional support relations. OC must be read jointly with coverage and navigability: sparse rooms can have excellent OC while failing the program and reachability tests.

\subsection{Controlled Same-Data Benchmark}
\label{sec:controlled_results}

\begin{center}
\centering
\captionof{table}{Controlled same-data benchmark on the shared 1{,}225-room geometry-valid subset. Cov. is coverage, SR is success rate, DF is detour factor, and OC is overlap--clearance. Higher is better for Cov., Nav, and SR; lower is better for DF and OC. Values are mean $\pm$ std; bold marks the best primary method per metric.}
\label{tab:controlled}
\setlength{\tabcolsep}{2pt}
\renewcommand{\arraystretch}{1.05}
\scriptsize
\begin{adjustbox}{width=\linewidth,center}
\begin{tabular}{lccccc}
\toprule
Method & Cov. $\uparrow$ & Nav $\uparrow$ & SR $\uparrow$ & DF $\downarrow$ & OC $\downarrow$ \\
\midrule
ATISS~\cite{Paschalidou2021NEURIPS} (same-data)             & 40.8 $\pm$ 23.6 & $-$30.3 $\pm$ 24.0 & 0.03 $\pm$ 0.18 & 0.97 $\pm$ 0.18 & \sotapm{0.5}{2.5} \\
BLT~\cite{10.1007/978-3-031-19790-1_29} (same-data)         & 17.8 $\pm$ 20.4 & $-$18.6 $\pm$ 42.7 & 0.12 $\pm$ 0.32 & 0.88 $\pm$ 0.31 & 35.7 $\pm$ 16.9 \\
\midrule
\sota{DDEP (Ours)}                                          & \sotapm{54.9}{38.3} & \sotapm{14.6}{56.0} & \sotapm{0.36}{0.42} & \sotapm{0.60}{0.44} & 13.4 $\pm$ 12.5 \\
\midrule
\multicolumn{6}{c}{\emph{DDEP ablations}}\\
\midrule
\quad Wall-Only                       & 50.8 $\pm$ 39.6 & 9.6 $\pm$ 54.8 & 0.32 $\pm$ 0.41 & 0.63 $\pm$ 0.44 & 12.7 $\pm$ 11.8 \\
\quad No Cont. Groups                 & 42.6 $\pm$ 40.2 & 0.7 $\pm$ 52.2 & 0.25 $\pm$ 0.39 & 0.69 $\pm$ 0.43 & 13.4 $\pm$ 13.8 \\
\quad No Edge Coords                  & 56.7 $\pm$ 39.7 & 14.4 $\pm$ 57.6 & 0.35 $\pm$ 0.43 & 0.60 $\pm$ 0.45 & 14.5 $\pm$ 12.7 \\
\quad Order: $t$ first                & 42.6 $\pm$ 39.1 & 4.1 $\pm$ 54.5 & 0.29 $\pm$ 0.41 & 0.70 $\pm$ 0.41 & 10.6 $\pm$ 12.0 \\
\quad Order: edge,$t$ desc.           & 46.5 $\pm$ 39.9 & 7.8 $\pm$ 54.8 & 0.31 $\pm$ 0.41 & 0.66 $\pm$ 0.42 & 13.4 $\pm$ 13.2 \\
\bottomrule
\end{tabular}
\end{adjustbox}
\end{center}

\Cref{tab:controlled} reports the same-data benchmark. DDEP attains the strongest joint functional profile among compared methods: $54.9$ coverage and $14.6$ navigability, beating ATISS by $+14.1$ coverage and $+44.9$ navigability, and BLT by $+37.1$ coverage and $+33.2$ navigability. It also achieves substantially better reachability (SR $0.36$ vs.\ $0.03$/$0.12$) and shorter detours (DF $0.60$ vs.\ $0.97$/$0.88$). ATISS's near-zero OC ($0.5\%$) reflects sparse outputs that avoid collisions while leaving almost all targets unreachable; BLT exhibits the opposite failure mode with $35.7\%$ OC. The key result is therefore not a single scalar win but a functional tradeoff: DDEP places enough program elements to satisfy the room while keeping paths reachable. Per-room inspection shows the strongest gains in bedrooms, living rooms, and master bedrooms; compact baths and laundries remain harder because small footprint errors quickly invalidate door clearance. ATISS targets 3D-FRONT-style scenes with axis-aligned bounding boxes and BLT targets 2D graphic-design layouts; we bridge their native outputs into our wall-referenced representation using the published descale formula plus an affine map to the physical room followed by nearest-wall projection. This bridge applies to both baselines but not to DDEP, which is native to the format, so we cannot rule out that part of the gap reflects conversion loss rather than modelling capacity. It is at least identical across both baselines and does not touch the ablation rows; a conversion-free comparison would require retraining them on wall-referenced targets.

\paragraph{Variance and model sharing}
The per-room deviations in \cref{tab:controlled} are large relative to the means because both metrics are bounded per room (Cov.\ in $[0,100]$, Nav in $[-35,100]$) with mass near the extremes: a room is typically either largely satisfied and traversable or largely blocked, and one obstructed doorway sends an otherwise sound room to the floor of the Nav scale. We therefore read Cov/Nav/OC as a joint profile rather than as individually significant scalars. The two protocols also bracket a design choice: production DDEP trains a $512$-dim specialist per room type on top of a pre-trained encoder, while the controlled model is a \emph{single} $256$-dim network trained from scratch on $15{,}740$ rooms and serving all five types. The shared, from-scratch setting is the weaker of the two, so \cref{tab:controlled} compares architectures in the harder regime. This brackets rather than measures the effect---capacity, pre-training, data, and test set all differ---so a fixed-capacity shared-versus-specialist ablation remains future work.

\paragraph{Ablations}
Five ablations isolate internal design choices within our tokenization. \emph{No Continuous Groups} ($-12.3$ cov, $-13.9$ nav) replaces the multi-continuous embedding with independent scalar MLPs and is the most damaging single change, confirming that joint embedding of correlated geometric features is the central inductive bias. \emph{Ordering ablations} (\emph{$t$-value First}: $-12.3$/$-10.5$; \emph{edge,\,$t$ Desc}: $-8.4$/$-6.8$) show that the canonical edge-then-position ordering is also material. \emph{Wall-Only} ($-4.1$/$-5.0$) confirms that opening positions help the decoder reason about clearances. \emph{No Edge Coordinates} ($+1.8$/$-0.2$) is nearly neutral on coverage and navigability while OC rises slightly, indicating the model infers relative geometry from remaining features when explicit endpoints are removed. Note that all five isolate \emph{within-tokenization} choices; the flat-serialization baseline that would test the structure itself is discussed in \cref{sec:conclusion}.

\subsection{Frontier LLM/VLM Baseline}
\label{sec:llm_floor}

\begin{table}[t]
\centering
\caption{Deployment-style held-out benchmark on 50 production rooms. Frontier rows report the best score achieved by any evaluated text-only LLM or VLM in each metric column; domain rows are individual systems. LayoutVLM requires a furniture list, so Coverage is not applicable. Bold marks the best result per metric.}
\label{tab:frontier}
\setlength{\tabcolsep}{3pt}
\renewcommand{\arraystretch}{1.02}
\begin{adjustbox}{width=\linewidth,center}
\begin{tabular}{lcccc}
\toprule
Method & Cov. $\uparrow$ & Nav. $\uparrow$ & OC $\downarrow$ & Lat. (s) $\downarrow$ \\
\midrule
\textbf{DDEP (ours)} & \textbf{98.1} & \textbf{79.2} & \textbf{1.9} & \textbf{3.18} \\
Text LLM envelope & 68.3 & 18.9 & 9.8 & 7--81 \\
VLM envelope & 65.6 & 55.5 & 8.2 & 7--102 \\
LayoutVLM~\cite{sun2025layoutvlm} & -- & 40.3 & 3.8 & 145 \\
FlairGPT~\cite{littlefair2025flairgpt} & 46.6 & 28.2 & 3.9 & 1134 \\
\bottomrule
\end{tabular}
\end{adjustbox}
\end{table}

\Cref{tab:frontier} summarizes the deployment-style benchmark. Production DDEP has the only high-coverage, high-navigability, low-OC profile: it reaches $98.1$ coverage, $79.2$ navigability, $1.9\%$ OC, and $3.18{\pm}0.18$ s latency. Frontier LLM/VLM systems close part of the semantic gap but remain less reliable spatially; domain-specific methods keep OC low only while missing large portions of the room program. To be precise about what the baselines received: every model gets the same tuned prompt with an explicit coordinate system, wall-condition guidance, a room-type-filtered entity catalog for information parity with DDEP, and a matched example of the expected output format, so the gap is not an artifact of naive prompting. It remains a calibration floor rather than an architectural ceiling because each model is run once per room with no best-of-$k$ or chain-of-thought, and no open model is fine-tuned in-domain---the informative next comparison.

\subsection{Embedding Evaluation}
\label{sec:embeddings}

We also evaluate SBM's encoder-only pathway against E5-Large-v2~\cite{wang2022e5}, BGE-Large-v1.5~\cite{bge_embedding}, and GTE-Large-v1.5~\cite{li2023towards} using serialized room text for the baselines. Text encoders lead on within-type nDCG (E5 $86.7$ vs.\ SBM $68.2$), but SBM produces substantially more coherent type-level clusters (NMI $0.726$ vs.\ $0.371$; ARI $0.437$ vs.\ ${\leq}0$), which is the property used by retrieval-augmented BIM generation.

\subsection{Qualitative Comparisons and Failure Modes}
\begin{figure}[t]
\centering
\includegraphics[width=0.92\linewidth]{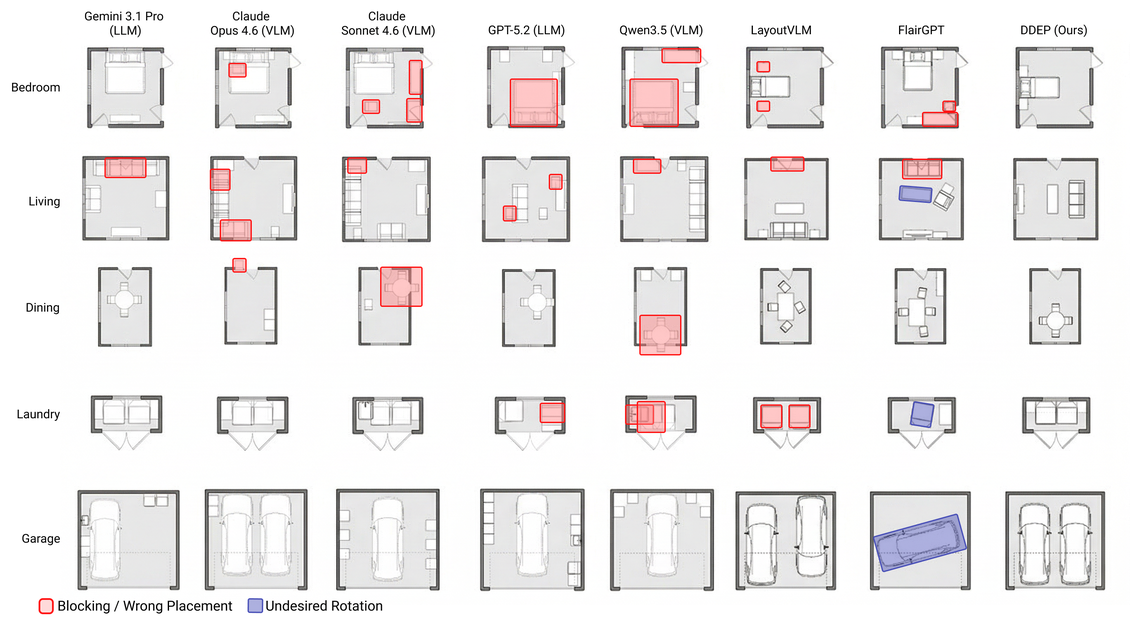}
\caption{2D diagnostic grid for representative room types. Red marks blocking or wrong placement; blue marks undesired rotation. DDEP avoids most highlighted failures while satisfying the room program.}
\label{fig:ddep_qual}
\end{figure}
\Cref{fig:hero} shows generated layouts across room types, and \cref{fig:ddep_qual} gives a 2D diagnostic view of the same comparison. VLM baselines tend to overfill rooms with extra casework and furniture, blocking circulation and door-swing zones. Domain-specific methods place more appropriate furniture types but still leave narrow or blocked paths around key elements. DDEP layouts satisfy the room program while maintaining clear, door-connected circulation bands and respecting wall-referenced anchoring. Characteristic DDEP failures include under-populating ambiguous open-plan rooms, dropping an entity when door clearance is over-tight, and misaligning casework groups on short wall segments.

\section{Conclusion}
\label{sec:conclusion}

We presented a BIM-native tokenization for room-level layout synthesis: sparse attribute--feature matrix columns for logical elements, wall-referenced coordinates, and a mixed-type embedding module for categorical, scalar, and grouped-continuous attributes. Reusing one Transformer for retrieval and DDEP, the method improves the controlled same-data benchmark against ATISS/BLT and produces stronger type-level clusters than text encoders. The core takeaway is representational: modest domain-specific sequence models over BIM-editable tokens can be an effective primitive for constraint-aware indoor layout synthesis, complementary to general-purpose LLMs/VLMs.

Several limitations bound these claims and set the agenda for future work. Our evaluation is limited to residential rooms from a proprietary corpus, which constrains external reproducibility; any future release of data or code will be announced on the project page. The LLM/VLM benchmark compares trained DDEP against zero-shot systems, so in-domain fine-tuning of an open model may close part of the gap. The ablations test \emph{within-tokenization} choices, which makes a flat-serialization baseline that removes the attribute--feature structure entirely the next critical experiment. We also omit direct HouseDiffusion~\cite{shabani2022housediffusion}/LayoutGPT~\cite{feng2023layoutgpt} comparisons because their polygonal or object--relation outputs need a non-trivial bridge to wall-hosted BIM entities. Non-residential typologies, building-scale layouts, vertical constraints, and local code variation remain open, as does disentangling model capacity from data scale in the shared-versus-specialist comparison.

\bibliographystyle{IEEEtran}
\bibliography{main}

\begin{thebibliography}{10}

\bibitem{eastman_spatial_1975}
C.~N. Eastman, {\em Spatial synthesis in computer-aided building design}.
\newblock Elsevier Science Inc., 1975.

\bibitem{MONEDERO2000369}
J.~Monedero, ``Parametric design: a review and some experiences,'' {\em
  Automation in Construction}, vol.~9, no.~4, pp.~369--377, 2000.

\bibitem{kan_automated_2017}
P.~Kán and H.~Kaufmann, ``Automated interior design using a genetic
  algorithm,'' in {\em Proceedings of the 23rd {ACM} {Symposium} on {Virtual}
  {Reality} {Software} and {Technology}}, {VRST} '17, (New York, NY, USA),
  pp.~1--10, Association for Computing Machinery, Nov. 2017.

\bibitem{Flager2007ACO}
F.~Flager and J.~R. Haymaker, ``A comparison of multidisciplinary design,
  analysis and optimization processes in the building construction and
  aerospace industries,'' in {\em A comparison of multidisciplinary design,
  analysis and optimization processes in the building construction and
  aerospace industries}, 2007.

\bibitem{merrell_interactive_2011}
P.~Merrell, E.~Schkufza, Z.~Li, M.~Agrawala, and V.~Koltun, ``Interactive
  furniture layout using interior design guidelines,'' {\em ACM Trans. Graph.},
  vol.~30, pp.~87:1--87:10, July 2011.

\bibitem{song_web3d-based_2019}
P.~Song, Y.~Zheng, J.~Jia, and Y.~Gao, ``{Web3D}-based automatic furniture
  layout system using recursive case-based reasoning and floor field,'' {\em
  Multimedia Tools and Applications}, vol.~78, pp.~5051--5079, Feb. 2019.

\bibitem{xu2018graph2seqgraphsequencelearning}
K.~Xu, L.~Wu, Z.~Wang, Y.~Feng, M.~Witbrock, and V.~Sheinin, ``Graph2seq: Graph
  to sequence learning with attention-based neural networks,'' 2018.

\bibitem{hu2020graph2plan}
R.~Hu, Z.~Huang, Y.~Tang, O.~van Kaick, H.~Zhang, and H.~Huang, ``Graph2plan:
  Learning floorplan generation from layout graphs,'' {\em arXiv preprint
  arXiv:2004.13204}, 2020.

\bibitem{tanasra_automation_2023}
H.~Tanasra, T.~Rott~Shaham, T.~Michaeli, G.~Austern, and S.~Barath,
  ``Automation in {Interior} {Space} {Planning}: {Utilizing} {Conditional}
  {Generative} {Adversarial} {Network} {Models} to {Create} {Furniture}
  {Layouts},'' {\em Buildings}, vol.~13, p.~1793, July 2023.
\newblock Publisher: Multidisciplinary Digital Publishing Institute.

\bibitem{liu_exploration_2024}
Y.~Liu and G.~Wang, ``Exploration of the {Indoor} {Layout} {Optimization}
  {Model} in {Computer}-{Aided} {Visual} {Analysis},'' {\em Computer-Aided
  Design and Applications}, pp.~167--180, Aug. 2024.

\bibitem{nauata2021housegan}
N.~Nauata, W.-C.~M. Chang, Y.~Furukawa, and et~al., ``House-gan++: Generative
  adversarial layout refinement network towards intelligent computational
  agent,'' in {\em CVPR}, 2021.

\bibitem{shabani2022housediffusion}
M.~A. Shabani, S.~Hosseini, and Y.~Furukawa, ``Housediffusion: Vector floorplan
  generation via a diffusion model with discrete and continuous denoising,''
  2022.

\bibitem{midiffusion2024}
S.~Hu {\em et~al.}, ``{MiDiffusion}: Mixed diffusion for 3d indoor scene
  synthesis,'' {\em arXiv preprint arXiv:2405.21066}, 2024.

\bibitem{semLayoutDiff2025}
X.~Sun {\em et~al.}, ``{SemLayoutDiff}: Semantic layout generation with
  diffusion models,'' {\em arXiv preprint arXiv:2508.18597}, 2025.

\bibitem{nguyen2024housecrafter}
H.~T. Nguyen, Y.~Chen, V.~Voleti, V.~Jampani, and H.~Jiang, ``Housecrafter:
  Lifting floorplans to 3d scenes with 2d diffusion model,'' in {\em arXiv
  preprint arXiv:2406.20077}, 2024.

\bibitem{layyourscene2025}
D.~Srivastava {\em et~al.}, ``Lay-your-scene: Natural scene layout generation
  with diffusion transformers,'' {\em arXiv preprint arXiv:2505.04718}, 2025.

\bibitem{directlayout2025}
X.~Ran {\em et~al.}, ``Directlayout: Direct numerical layout generation for 3d
  indoor scene synthesis,'' {\em arXiv preprint arXiv:2506.05341}, 2025.

\bibitem{Paschalidou2021NEURIPS}
D.~Paschalidou, A.~Kar, M.~Shugrina, K.~Kreis, A.~Geiger, and S.~Fidler,
  ``Atiss: Autoregressive transformers for indoor scene synthesis,'' in {\em
  Advances in Neural Information Processing Systems (NeurIPS)}, 2021.

\bibitem{wang2020sceneformer}
X.~Wang, C.~Yeshwanth, and M.~Nie{\ss}ner, ``Sceneformer: Indoor scene
  generation with transformers,'' {\em arXiv preprint arXiv:2012.09793}, 2020.

\bibitem{Liu2023CLIPLayoutSI}
J.~Liu, W.~Xiong, I.~Jones, Y.~Nie, A.~Gupta, and B.~Ouguz, ``Clip-layout:
  Style-consistent indoor scene synthesis with semantic furniture embedding,''
  {\em ArXiv}, vol.~abs/2303.03565, 2023.

\bibitem{10.1007/978-3-031-19790-1_29}
X.~Kong, L.~Jiang, H.~Chang, H.~Zhang, Y.~Hao, H.~Gong, and I.~Essa, ``Blt:
  Bidirectional layout transformer for controllable layout generation,'' in
  {\em Computer Vision – ECCV 2022: 17th European Conference, Tel Aviv,
  Israel, October 23–27, 2022, Proceedings, Part XVII}, (Berlin, Heidelberg),
  p.~474–490, Springer-Verlag, 2022.

\bibitem{feng2023layoutgpt}
W.~Feng, W.~Zhu, T.-J. Fu, V.~Jampani, A.~R. Akula, X.~He, S.~Basu, X.~E. Wang,
  and W.~Y. Wang, ``Layoutgpt: Compositional visual planning and generation
  with large language models,'' in {\em Advances in Neural Information
  Processing Systems}, 2023.

\bibitem{sun2025layoutvlm}
F.-Y. Sun, W.~Liu, S.~Gu, D.~Lim, G.~Bhat, F.~Tombari, M.~Li, N.~Haber, and
  J.~Wu, ``Layoutvlm: Differentiable optimization of 3d layout via
  vision-language models,'' in {\em Proceedings of the IEEE/CVF Conference on
  Computer Vision and Pattern Recognition (CVPR)}, pp.~29469--29478, 2025.

\bibitem{zong2024housetune}
Z.~Zong, G.~Chen, Z.~Zhan, F.~Yu, and G.~Tan, ``Housetune: Two-stage floorplan
  generation with {LLM} assistance,'' 2024.

\bibitem{littlefair2025flairgpt}
G.~Littlefair, N.~S. Dutt, and N.~J. Mitra, ``Flairgpt: Repurposing llms for
  interior designs,'' 2025.
\newblock EUROGRAPHICS 2025.

\bibitem{li2024llm4cad}
X.~Li, Y.~Sun, and Z.~Sha, ``Llm4cad: {Multi-Modal} large language models for
  three-dimensional computer-aided design generation,'' in {\em Proceedings of
  the ASME 2024 International Design Engineering Technical Conferences and
  Computers and Information in Engineering Conference (IDETC/CIE 2024)},
  vol.~88407, p.~V006T06A015, ASME, 2024.

\bibitem{yin2025floorplandeepseek}
J.~Yin, P.~Zeng, J.~Zhong, P.~Li, M.~Zhang, R.~Luo, and S.~Lu,
  ``Floorplan-deepseek (fpds): A multimodal approach to floorplan generation
  using vector-based next room prediction,'' {\em arXiv preprint},
  vol.~arXiv:2506.21562, 2025.

\bibitem{wang2022e5}
L.~Wang, N.~Yang, X.~Huang, B.~Jiao, L.~Yang, D.~Jiang, R.~Majumder, and
  F.~Wei, ``Text embeddings by weakly-supervised contrastive pre-training,''
  {\em arXiv preprint arXiv:2212.03533}, 2022.

\bibitem{bge_embedding}
S.~Xiao, Z.~Liu, P.~Zhang, and N.~Muennighoff, ``C-pack: Packaged resources to
  advance general chinese embedding,'' 2023.

\bibitem{li2023towards}
Z.~Li, X.~Zhang, Y.~Zhang, D.~Long, P.~Xie, and M.~Zhang, ``Towards general
  text embeddings with multi-stage contrastive learning,'' {\em arXiv preprint
  arXiv:2308.03281}, 2023.

\end{thebibliography}

\end{document}